\documentclass{llncs}
\pdfoutput=1

\usepackage{amsmath}
\usepackage{graphicx}
\usepackage{xcolor}
\usepackage{array}
\newcolumntype{P}[1]{>{\centering\arraybackslash}p{#1}}

\begin{document}
\title{Identifying Points of Interest and Similar Individuals from Raw GPS Data}
\titlerunning{Mobility IoT 2018 - 5th EAI International Conference on Smart Cities within SmartCity\textdegree360 Summit}
%
\author{Thiago Andrade\inst{1} \and
Jo\~ao Gama\inst{1,2}}


%
\institute{INESC TEC, Porto, Portugal 
\and
University of Porto, Porto, Portugal\\
\email{thiago.a.silva@inesctec.pt}\\
\email{jgama@fep.up.pt}
}

\maketitle              
\begin{abstract}
Smartphones and portable devices have become ubiquitous and part of everyone's life. Due to the fact of its portability, these devices are perfect to record individuals' traces and life-logging generating vast amounts of data at low costs. These data is emerging as a new source for studies in human mobility patterns raising the number of research projects and techniques aiming to analyze and retrieve useful information from it. The aim of this paper is to explore GPS raw data from different individuals in a community and apply data mining algorithms to identify meaningful places in a region and describe user's profiles and its similarities. We evaluate the proposed method with a real-world dataset. The experimental results show that the steps performed to identify points of interest (POIs) and further the similarity between the users are quite satisfactory serving as a supplement for urban planning and social networks.


\end{abstract}
\section{Introduction}
\label{sec:Introduction}

Smartphones and portable devices have become ubiquitous and a part of everyone's life. Due to the fact of its portability, those devices acting as sensors are perfect to record individuals' traces and life-logging generating vast amounts of data at low costs. The popularity of the Location-based social networks (LBSN) has increased and raised curiosity and interest among Internet users which are more and more interacting with the many available services like Instagram, Facebook, Flickr. When dealing with raw data, final users cannot make sense of it without processing and apply techniques to extract meaningful information from its content. Many researchers have made efforts in exploring these data in order to find places, locations, and regions \cite{li2008mining,zheng2009mining,zheng2010geolife,cao2010mining,lee2013mining}. Hence, individuals can state a place as something with a meaning such as work, home, university while a pair like 30.2319, 120.14785 has no useful meaning to them.

Finding similar users is one of the most important tasks when dealing with social network services since one of the main issues is recommending similar users to a new user or even locations, products and events. In this sense, some authors have been working to find and propose methods to address this cause. The approach we used in this paper is based on the users' preferred points. 

In this work we tried to answer the following questions:
\textit{
\begin{itemize}
\item Is it possible to identify from raw GPS history data the users and community locations? 
\item Among those places, which ones are the points and regions of interest? 
\item Which are the users that share the most common places?
\end{itemize}
}

Here we present an approach to analyzing users' raw data looking for points of interest in a given community. We do this by processing individuals' trajectories in order to find stay points (SP) (locations where the user spend considerable time in a given radius). Then we transform the users' found stay points into individuals' location points (LP) by applying a density-based clustering algorithm. Finally, we perform density-based clustering in all users' location points (LP) of a given community in order to find the points of interest (POI). Another important item of this paper is the method to find similar users based on those points of interest of the community. All the above-mentioned tasks were evaluated in a real-world dataset.

The following section presents the literature review and the most important related works. The remainder of the paper describes the methodology and the data set in section 3, in section 4 we discuss the experiments and results obtained. Finally, the conclusions and future work are presented in section 5.

\section{Related Work} 
\label{sec:RelatedWork}

Many researchers have been proposing methods to extract points of interest and calculate the similarity between users for diverse goals. The recommendation of users is one of the most popular among them.
According to \cite{cao2010mining,lee2013mining}, several methods based on density have been proposed in order to discover regions of interest although most of these methods are used to aggregate spatial point objects.
Some authors were more interested in the semantic movement trajectories. \cite{li2008mining} introduced a model that makes use of movement datasets which has trajectories defined as sequences of timestamped stops and moves between locations. These trajectories are enriched with semantic meanings. From this, they define dense regions, where many different users had stop as regions of interest (ROI)
The social matching framework was proposed by \cite{terveen2005social} with the objective of match people by using their physical locations.

In order to help users on planning a trip to unknown places, \cite{lu2010photo2trip} presented a framework that makes use images (e.g., Flickr Images) and textual travelogues to perform recommendations.
\cite{de2010automatic} proposed something similar also making use of Flickr tagged images to automatically suggest and construct travel itineraries.
Another approach is proposed by \cite{zheng2009mining,zheng2010geolife} which uses the Geolife GPS logs to understand the relations of trajectories, users and the locations where they passed through to support the travelers to plan their trips. The authors make use of machine learning to build a recommender system by mining high ranked locations. 
\cite{zheng2009geolife2} performed a study based on the visiting pattern of users in distinct locations to define a similarity metric which detects similar users and their groups;

\section{Methodology}
\label{sec:Methodology}

Before entering in details of the methodology, we introduce the definition of a trajectory which will be used along the paper:\\

	\textit{Definition 1:} Trajectory: a trajectory is a list of ordered GPS points,\\ 
    $T \textit{=} {P0, P1, . . . , Pn}$, where $t0 < t1 < . . . < tn$ and $i = 0, 1, . . . , n$. \\

In this work, we denote a new trajectory every time an individual stop moving for more than 30 minutes.


The first step is the preprocessing task which is including among other activities, the data cleaning process where we perform outliers and noise removal. With the processed dataset we move to the second step where we process all the users' trajectories in order to find the stay points (SP). After having the set of stay points of the individuals' trajectories, we apply the algorithm to find the location points (LP). Further, after processing all users' location points, we perform the step responsible to extract the points of interest (POI) in the given community of users by using density-based clustering methods. And the last step of the proposed approach is to calculate the similarity between the individuals of the community based on the common POIs each pair of users have visited.

Following we describe the real-world dataset used to apply the proposed method.

\subsection{Dataset}
\label{subsec:Dataset}
All the activities in this list were conducted over the Geolife dataset. This GPS trajectory dataset was collected in (Microsoft Research Asia) Geolife project by 182 users in a period of over three years (from April 2007 to August 2012). A GPS trajectory of this dataset is represented by a sequence of time-stamped points, each of which contains the information of latitude, longitude, and altitude. This dataset contains 17.621 trajectories with a total distance of about 1.2 million kilometers and a total duration of 48.000+ hours. These trajectories were recorded by different GPS loggers and GPS-phones, and have a variety of sampling rates. 91\% of the trajectories are logged in a dense representation, e.g. every 1 to 5 seconds or every 5 to 10 meters per point. This dataset recorded a broad range of users' outdoor movements, including not only life routines like go home and go to work but also some entertainments and sports activities, such as shopping, sightseeing, dining, hiking, and cycling \cite{zheng2008understanding,zheng2009mining,zheng2010geolife}.

As we stated in section \ref{sec:Introduction}, this work makes use of clustering algorithms based on the density of the data. Density-based methods require data to be collected at more frequent intervals. As mentioned above, more than 90\% of the GPS communication intervals in the data set are less than 10$s$. This interval is adequate for the application of density-based methods to distinguish stationary points and regions of interest.

\subsection{Preprocessing}
\label{subsec:Preprocessing}

When analyzing sources of information and comparing GPS signals with others, one can say that GSM cell tower has an advantage on being available in indoors while GPS signals are not. In addition, GPS data often suffer from the called signal shadowing, when a given sensor is found inside vehicles, behind trees or buildings.  On the other hand, GSM cell tower signals give us a more coarse and imprecise register of the location.

Because of the influence of GPS signal loss and data drift, there are a number of outliers in the trajectory data during the data acquisition. 
Hence, cleaning tasks need to be performed in order to have more trustworthy data. In the given scenario of locating meaningful places, another relevant role of preprocessing is to avoid peaks to fall into other clusters nearby and may form a new stay point unnecessarily.

Examples of common situations are individuals that suddenly took a very high speed in a very short period of time, what is quite improbable. To remove this type of noise we apply a smoothing filter to each pair of GPS points (p1, p0) of the trajectory. 

\begin{figure}[!h]
\centering
     \includegraphics[width=0.48\textwidth]{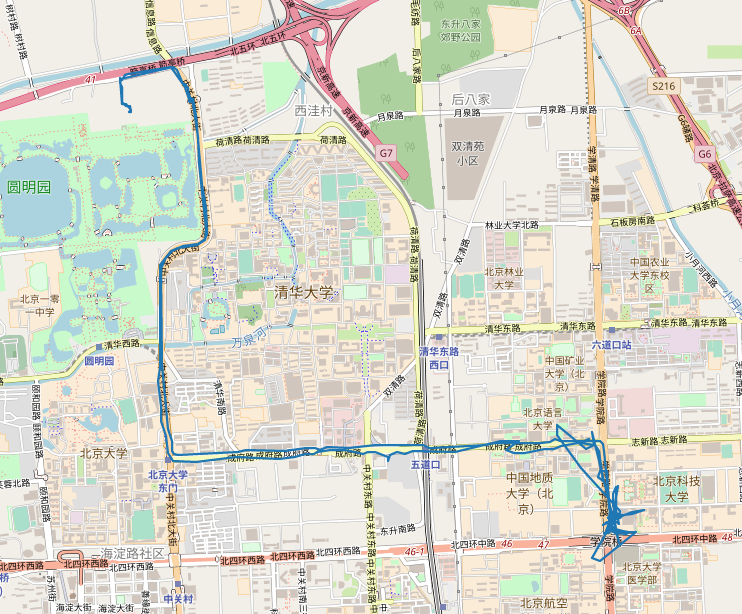}
    \includegraphics[width=0.488\textwidth]{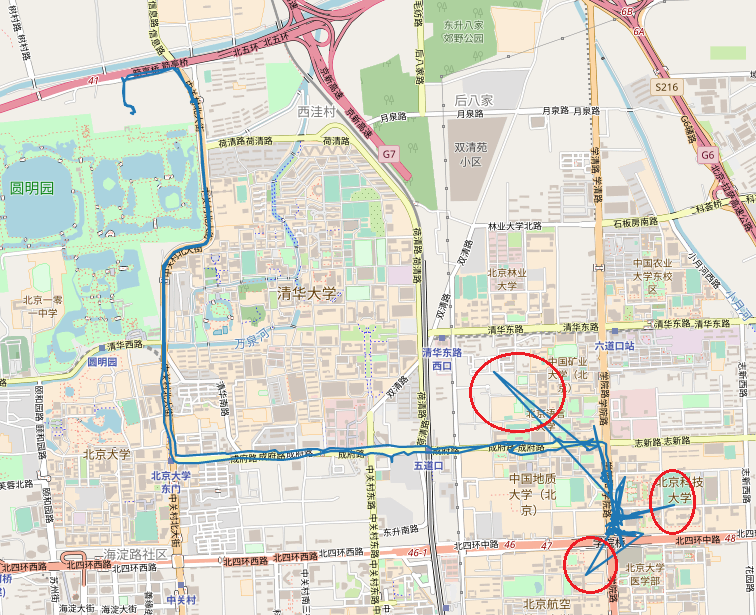}
\caption{Data cleaning process removing outliers and spikes in trajectory data. On the right image, we have the red ellipsis highlighting the spikes in the trajectory} \label{fig:spikesInGPSsignal}
\end{figure}




\subsection{User Stay Points Detection}
\label{subsec:UserStayPointsDetection}

Stay points are regions where a given user has stayed for a while within a defined radius. The algorithm used to extract the users' stay points from a trajectory is a hybrid density and time-based proposal \cite{ye2009mining}. It follows the logic of calculations between consecutive points where we measure the Haversine distance between two sets of coordinates p1 and p0 in order to find those that are below a distance threshold parameter. While the distance between the successive points remains smaller than the parameter, we keep adding them to a list of candidate points to form the stay point region. Next, the algorithm checks for how long the user stayed in that radius of the distance threshold checking for the second parameter, the time threshold. If the spent time is greater than the parameter value, we add those points to the final set of items which will form the new stay point.

\begin{figure}[!h]
\centering
\includegraphics[width=0.8\textwidth]{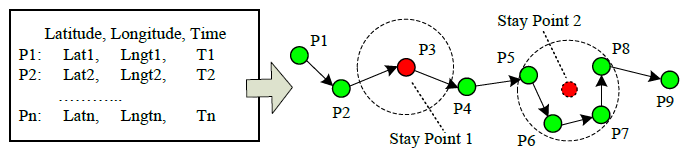}
\caption{GPS log and stay points \cite{ye2009mining}} \label{fig:stayPointsDef}
\end{figure}

For this experiment, we set the parameters Distance-threshold as 200 meters and the Time-threshold to 20 minutes. Hence, individuals having in their trajectories, consecutive points in a region of 200 meters radius for more than 20 minutes are creating new stay points for each similar situation. Having a new set of points we need to calculate their centroid which we denote as the center of mass of all the points in the cluster. To perform this task we calculate the mean of the coordinates of the set of points in the cluster.
We can see an example of stay point in the Figure \ref{fig:stayPointsDef}

\subsection{User Location Points}
\label{subsec:UserLocationPoints}

A location point is defined as a frequent location visited by an individual. For the purpose of finding points of interest which are described in section \ref{subsec:PointsOfInterest}, we need to find the most common locations within all the users in the community individually. In this way, we look for those places a person visits repeatedly in order to form the so-called users' location points. 

Location detection techniques are common tasks which make use of density-based methods, this is due because the mechanism of density-based clustering enables it to detect clusters of arbitrary shapes without specifying the number of clusters in the data a priori. It also has a notion of noise and is tolerant of outliers.

\begin{figure}[!h]
\centering
\includegraphics[width=0.6\textwidth]{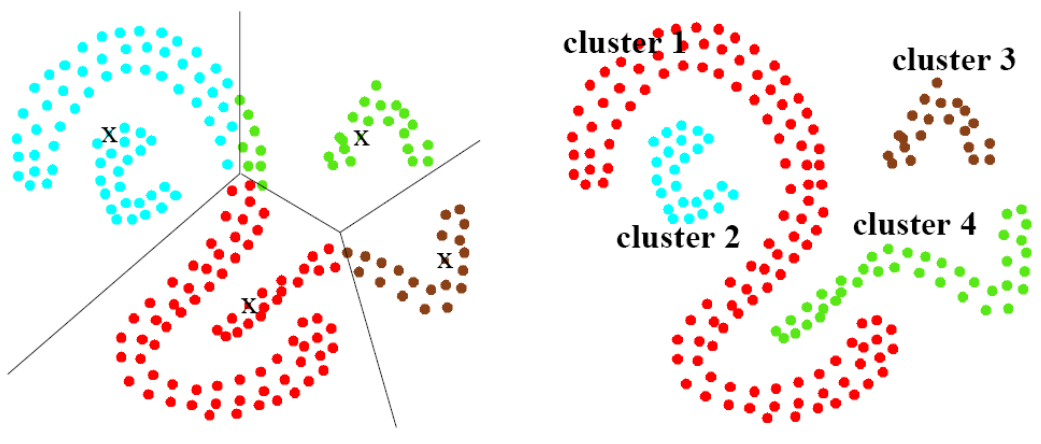}
\caption{A comparison between clustering based in centroid distance and density based cluster} \label{fig:clusters_differences}
\end{figure}

In this way, the clustering algorithm we selected to perform this task is DBSCAN (Density-Based Spatial Clustering of Applications with Noise) \cite{ester1996density}. Among other advantages, and because of the fuzziness of the trajectory data points logging, this algorithm performs very well with geographic data with arbitrary shape, are easily adaptive to different distance functions and also are able to detect noise which is very useful in our case as we do not want points that not fall close to each other to be part of our points of interest set. In figure \ref{fig:clusters_differences} we can verify a comparison between density based and regular centroid distance cluster.

The parameters used in the algorithms' set up are the Eps, related to the distance between the points and the MinPts which are the number of minimum points required to form a cluster. Some techniques were proposed to estimate these parameters, in this work we are following \cite{ester1996density} k-dist heuristic to determine their optimal values. After applying the heuristic we ended up with the values 100 meters for the Eps (distance) and 4 points to the MinPts (minimum points to form a cluster)

\subsection{Points Of Interest}
\label{subsec:PointsOfInterest}

Aiming to extract the most common locations among all users in a given community, a step of clustering the user location points (LP) extracted in section \ref{subsec:UserLocationPoints}, finding those Points of Interest visited by many individuals. The parameters utilized to perform this task are slightly different than those used to extract the location points as to form points of interest in a community the Eps (distance) and MinPts should be increased. The experiments are shown in section \ref{sec:ExperimentsResults}


\subsection{Users Similarity}
\label{subsec:UsersSimilarity}

By having the set of points of interest of all users in the community, we can apply the method to calculate the users' similarity by verifying the number of points of interest the individuals share in common.
The similarity measure utilized in this method is the Jaccard’s Coefficient described in the equation \ref{eq:Jaccard}. This coefficient is denoted by the number of locations shared by two users divided by the size of their locations’ union, characterizing the similarity between their sets of locations. 
In this way, we calculate the users' intersection set which are the locations shared by the two individuals and calculate the union of all locations visited by them. Hence be A the set of locations visited by a user u1 and B the set of locations visited by a user u2, the intersection set (A $\cap$ B) and the union set (A $\cup$ B), for both users, are respectively the set of locations that both individuals have visited and the set of regions visited for at least one of them.
	
\begin{equation}\label{eq:Jaccard}
	Jaccard(x, y) \textit{=}  \frac{|\Gamma (x) \cap \Gamma (y)|}{|\Gamma (x) \cup \Gamma (y)|} 
\end{equation}

\begin{figure}[!h]
\centering
\includegraphics[width=0.5\textwidth]{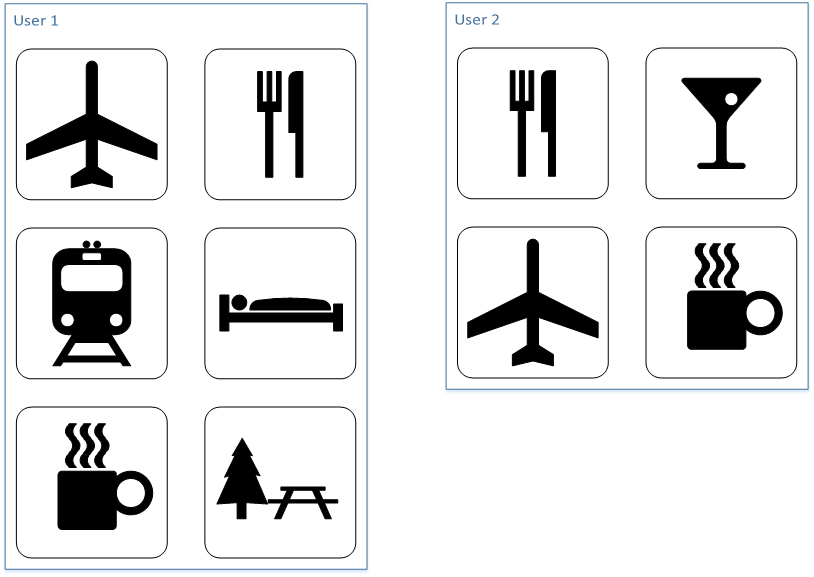}
\caption{Set of visited POIs of two given users} \label{fig:simlilarUsersPOIs}
\end{figure}

In Figure \ref{fig:simlilarUsersPOIs} we have an example of a set of POIs that a pair of users (A and B) have visited. In this particular scenario, the coefficient given by the similarity formula is represented as:\\ 

\begin{center} 
$|A \cup B| \textit{=} \begin{bmatrix} 01 & 02 & 03\\ 04 & 05 & 06 \end{bmatrix}$ \textit{=} 6 \; 
$|A \cap B| \textit{=} \begin{bmatrix} 02 & 07 & 01\\ 05 &    &    \end{bmatrix}$ \textit{=} 4 \;

$Sim(u1, u2) = \frac{3}{7}$
\end{center}

\section{Experiments and Results}
\label{sec:ExperimentsResults}

To perform the experiments, a group of 10 individuals representing a community was selected from the whole Geolife dataset in order to evaluate the proposed methods. The subset used to perform the evaluation shown in table \ref{tabExperimentsSubSet}

\begin{table}[!h]
\centering
\caption{Example of Geolife dataset.}\label{tabExperimentsSubSet}
\begin{tabular}{|P{4.5cm}|P{5.5cm}|}
\hline 
{\bfseries Period} &  {\centering April 2007 to August 2012} \\
\hline 
{\bfseries Users} &  {\centering 10} \\
\hline 
{\bfseries Number of trajectories} & {\centering 3.609} \\
\hline 
{\bfseries Number of GPS points} & {\centering 1.844.250} \\
\hline
\end{tabular}
\end{table}


\subsection{User Stay Points Detection - Results}
\label{subsec:UserStayPointsDetectionResults}

\begin{figure}[!h]
\centering
\includegraphics[width=0.6\textwidth]{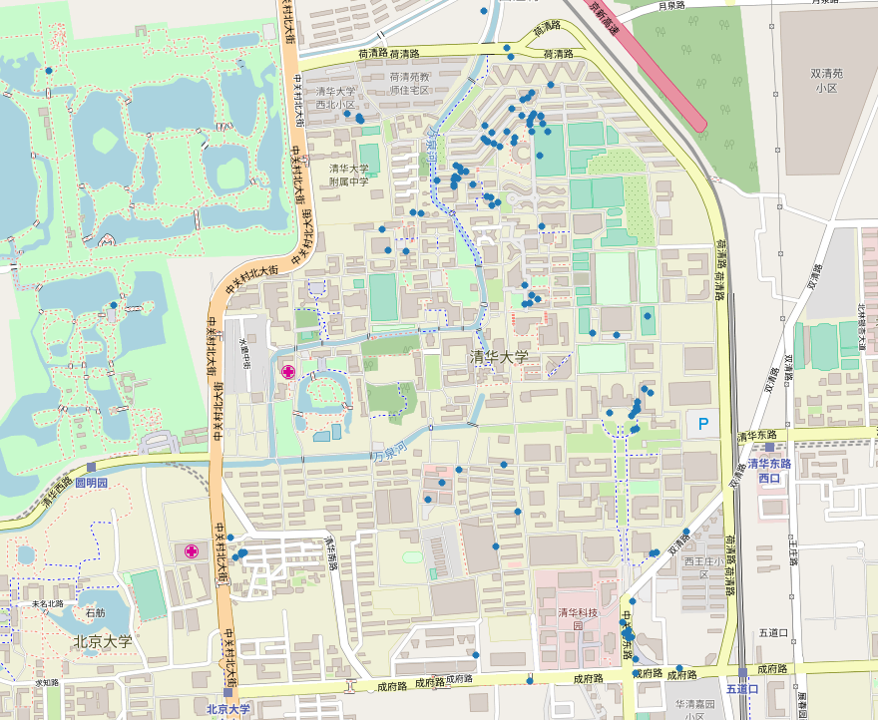}
\caption{Example of stay points (SP) found in the subset of GPS raw data by performing the proposed method over the user $'000'$. Each blue dot stands for a distinct stay point.} \label{fig:userStayPoints}
\end{figure}

To perform the search for stay points (SP), we set the algorithm parameters as following: radius distance threshold was set to 200 meters and the time threshold was set to 20 minutes. After running the stay points detection method, we were able to identify various points along the city. Figure \ref{fig:userStayPoints} shows examples of stay points found for the user $'000'$ in the subset.

In table \ref{tabTop5StayPoints} we summarize the top 10 users ranked by their number of stay points found  and the number of GPS trajectory points.

\begin{table}[!h]
\centering
\caption{Top 10 ranked users according to quantity of stay points detected and number of GPS points}\label{tabTop5StayPoints}
\begin{tabular}{|P{2.5cm}|P{2.5cm}|P{2.5cm}|P{2.5cm}|}
\hline
\textbf{User} & \textbf{\begin{tabular}[c]{@{}c@{}}Number of\\ Trajectories\end{tabular}} & \textbf{\begin{tabular}[c]{@{}c@{}}Number of\\ GPS Points\end{tabular}} & \textbf{\begin{tabular}[c]{@{}c@{}}Number of \\ Stay Points\end{tabular}} \\ 
\hline
004           & 1.217                                                                      & 439.397                                                                  &         2.671                                                                  \\ \hline
003           & 941                                                                       & 485.226                                                                  &           2.244                                                                \\ \hline
000           & 458                                                                       & 173.870                                                                  &                 2.098                                                          \\ \hline
002           & 329                                                                       & 248.217                                                                  &        786                                                                   \\ \hline
005           & 164                                                                       & 109.046                                                                  &          457                                                                 \\ \hline
001           & 130                                                                       & 108.607                                                                  &     438                                                                      \\ \hline
007           & 134                                                                       & 85.531                                                                   &       351                                                                    \\ \hline
009           & 109                                                                       & 84.616                                                                   &        314                                                                   \\ \hline
008           & 71                                                                        & 77.910                                                                   &        176                                                                   \\ \hline
006           & 56                                                                        & 31.830                                                                   &         116                                                                  \\ \hline
\hline
\textbf{Total}           & \textbf{3.609}                                                                       & \textbf{1.844.250}                                                                  &  \textbf{9.651}                                                                         \\ \hline
\end{tabular}
\end{table}


\subsection{User Location Points - Results}
\label{subsec:UserLocationPointsResults}

\begin{figure}[!h]
\centering
\includegraphics[width=0.9\textwidth]{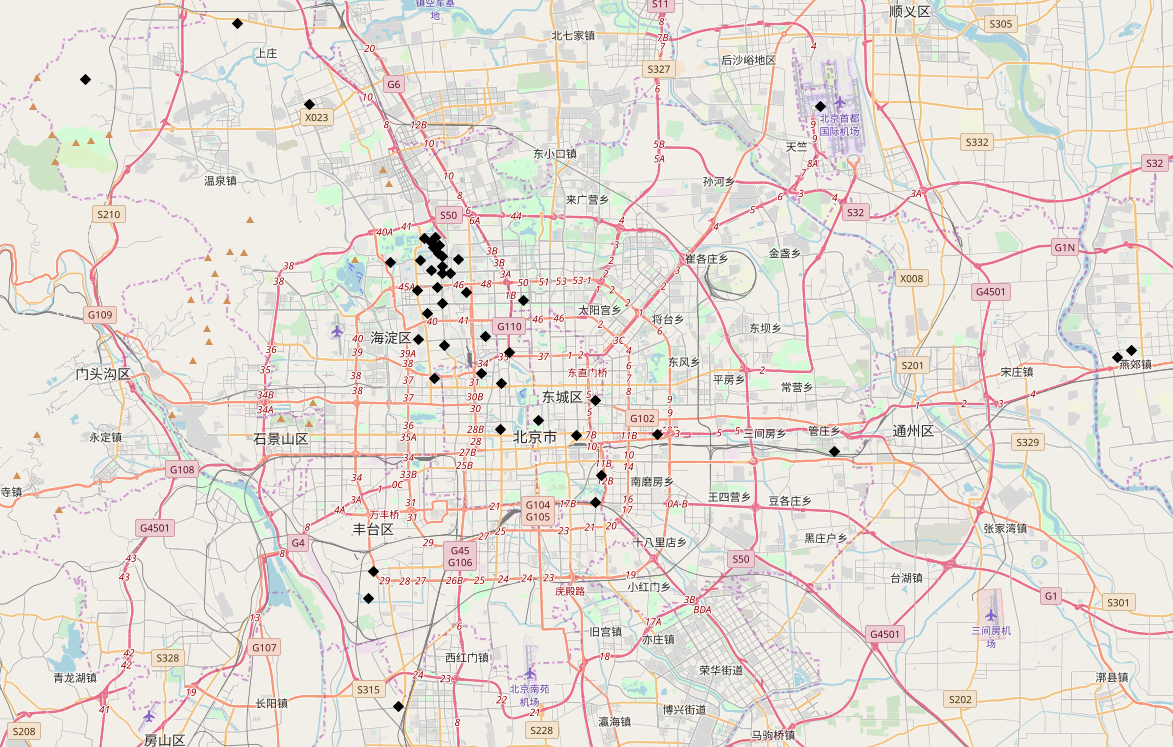}
\caption{Location points (LP) identified from the set of stay points (SP) of users in the subset in great Beijing area. Each black dot stands for a distinct location point. The majority of the location points of this given user are concentrated in the northwest area}
\label{fig:userLocationPoints}
\end{figure}

Following the framework pipeline, the next step was to identify the users' location points (LP). As mentioned in section \ref{subsec:UserLocationPoints}, a location point means a person frequently visits the location. 
In this study, we took care of filtering the locations visited occasionally by the user, as these locations do not represent meaningful places but some odd situations in users' life.
To solve this issue we use the location points in order to list the top-k locations.
The top-k locations are found by the most visited locations in users' history. When talking about mobility patterns, there are basically two groups of users, those who follow routines, visiting a few locations more frequently and those who use to explore more the region. 
We perform density-based cluster with a distance threshold (eps) of 100 meters and a minimum number required to form a cluster (MinPts) of 4 points, which means an individual that visits the same group of stay points that are located within a shorter radius of 100 meters at least 4 times consider that place as a meaningful location. Figure \ref{fig:userLocationPoints} show the user location points found in the experiments.


The set of stay points per user which were converted into location points can be verified in table \ref{tabTop5LocationPoints}

\begin{table}[!h]
\centering
\caption{Top 10 users' stay points converted into location points}\label{tabTop5LocationPoints}
\begin{tabular}{|P{1.5cm}|P{2.5cm}|P{2.5cm}|P{7.0cm}|}
\hline
\textbf{\begin{tabular}[c]{@{}c@{}}User\end{tabular}}  & \textbf{\begin{tabular}[c]{@{}c@{}}Number of \\ Stay Points\end{tabular}} & \textbf{\begin{tabular}[c]{@{}c@{}}Number of \\ Location Points \end{tabular}} \\ \hline
004            &         2.671                                           &         41                                              \\ \hline
003            &           2.244                                         &       46                                                \\ \hline
000            &        2.098                                            &         38                                              \\ \hline
002            &         786                                           &          23                                             \\ \hline
005            &         457                                           &             5                                          \\ \hline
001            &         438                                           &           18                                            \\ \hline
007            &        351                                            &            8                                           \\ \hline
009            &        314                                            &            4                                           \\ \hline
008            &        176                                            &            7                                           \\ \hline
006            &        116                                            &             2                                          \\ \hline
\hline
\textbf{Total} & \textbf{9.651}                                          & \textbf{192}                                             \\ \hline
\end{tabular}
\end{table}


\subsection{Points Of Interest - Results}
\label{subsec:PointsOfInterestResults}

\begin{figure}[!h]
\centering
\includegraphics[width=0.9\textwidth]{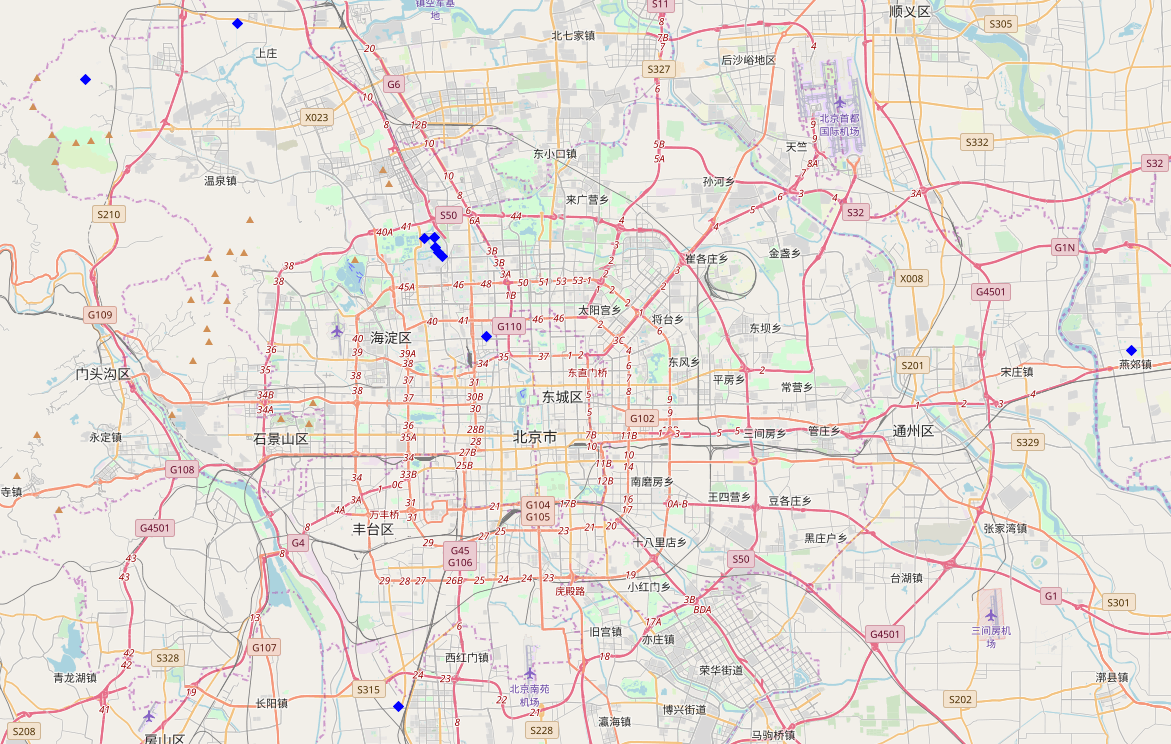}
\caption{Points of interest (POI) identified from the set of all users' location points (LP) of the community in great Beijing area. Although there are POIs all over the city, most of them are concentrated in the northwest area} 
\label{fig:POIS}
\end{figure}

The points of interest (POI) concept is related to small regions where a considerable amount of visits have occurred. To extract the points of interest in the given community of users we need to process the individual's location points found in subsection \ref{subsec:UserLocationPointsResults} in order to identify those locations with a higher number of visits among all the dataset. We achieve this by performing density-based clustering in those location points configuring the distance (eps) parameter to 200 meters and the minimum points (minPts) to 4. In this way, only those location points that have been visited in at least 4 occasions are elected as points of interest. This generated 32 POIs found in the selected group of users for this community. The results of the process are in Figure \ref{fig:POIS}.


\subsection{Users Similarity - Results}
\label{subsec:UsersSimilarityResults}

Regarding the results about the users' similarity is important to notice that by using individuals' location GPS logs history, we cannot identify the real intention the users have been to a given place as density-based algorithms are supported by a given radius of distance in order to aggregate the closest points. Whereas in the real world, one specific physical location can be related to more than one place, especially when dealing with building with levels such as shopping centers, hospitals or universities. 

The results obtained regarding users' similarities are listed in table \ref{tabUsersSimilarity}. The data is ranked from the most similar users to the less similar ones according to the proposed Jaccard's coefficient mentioned in section \ref{subsec:UsersSimilarity}

\begin{table}[!h]
\centering
\caption{Top 5 similar users found in the community according to Jaccard similarity}\label{tabUsersSimilarity}
\begin{tabular}{|c|c|c|c|}
\hline
\textbf{User A} & \textbf{User B} & \textbf{Similar places} & \textbf{Similarity percentage} \\ \hline
000               & 003               & 9/10                & 0.90                \\ \hline
000               & 004               & 8/9                 & 0.88                \\ \hline
003               & 004               & 8/10                & 0.80                \\ \hline
005               & 008               & 1/3                 & 0.33                \\ \hline
003               & 005               & 3/10                & 0.30                \\ \hline
\end{tabular}
\end{table}

\begin{figure}[!h]
\centering
     \includegraphics[width=0.48\textwidth]{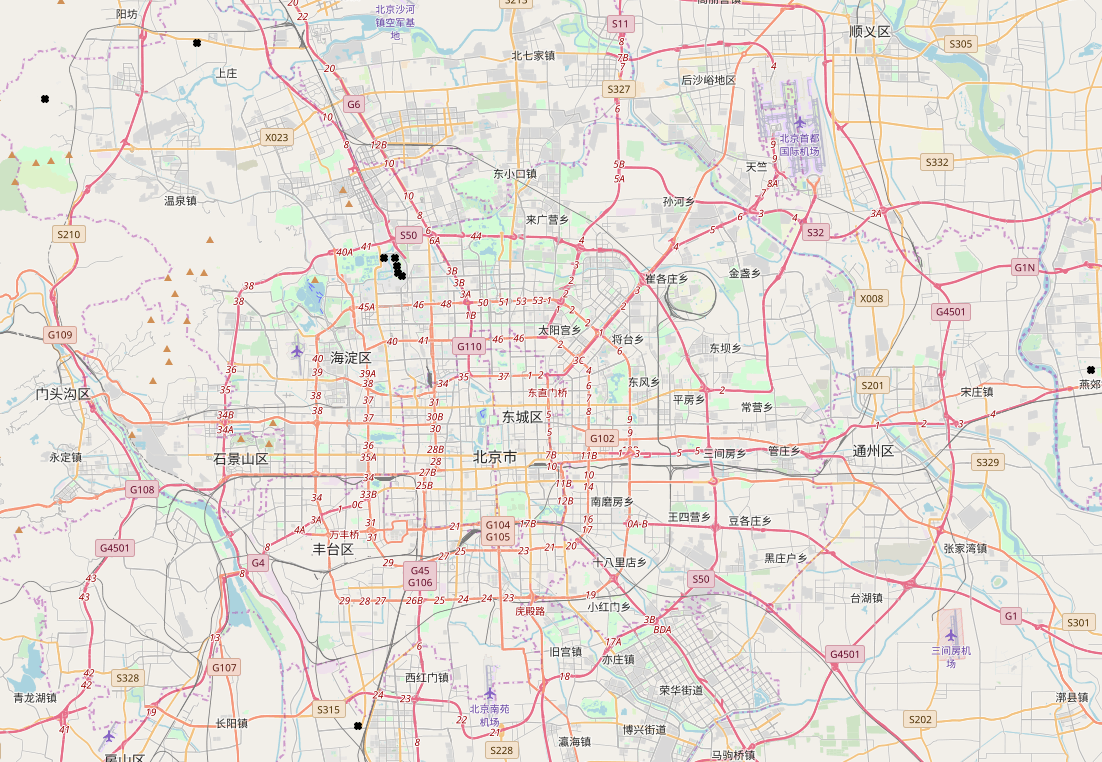}
    \includegraphics[width=0.49\textwidth]{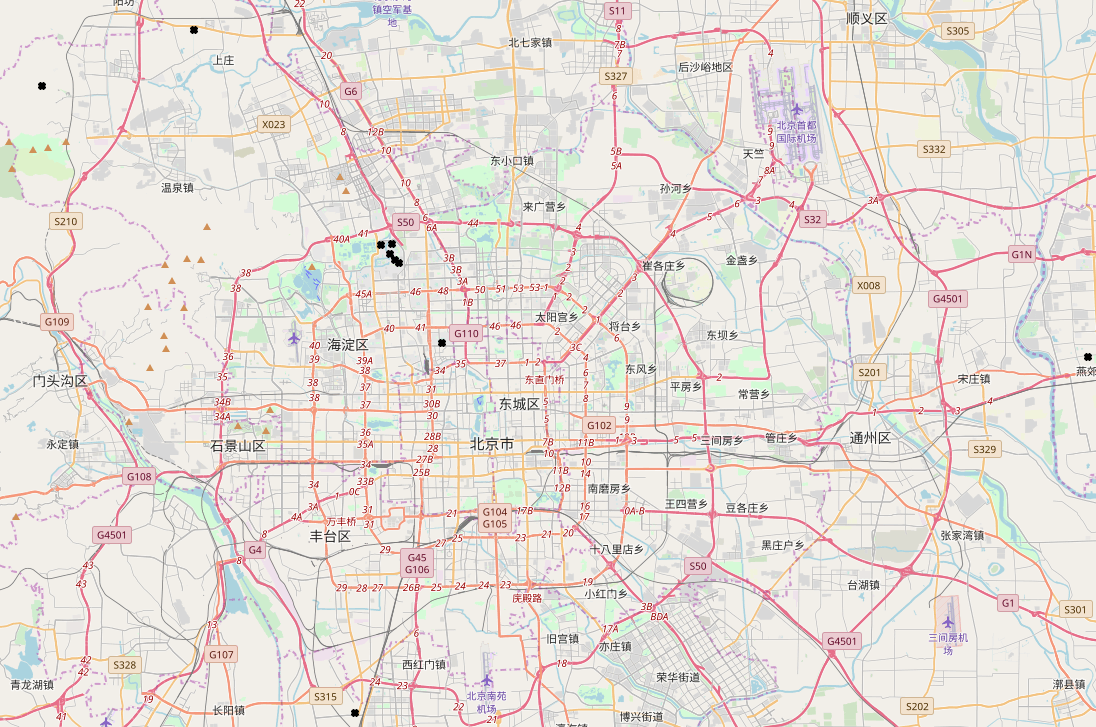}
\caption{Similarity between a pair of users that are top ranked in the similarity index results. The black dots represent the common POIs shared.} \label{fig:UsersSimilarity}
\end{figure}

In order to analyze the similarity results in a geographical manner, we have the Figure \ref{fig:UsersSimilarity} representing the POIs of the two most similar users in the community. One can notice the closeness of their locations having 9 out of 10 locations in common corresponding to 90\% of similarity. 

The highest the degree, the more likely to be potential friends or have similar location preferences. The results of this simple method can be useful to understand the dynamics and relations between users in a given community. Possible applications of this work are in recommendation tasks for locations, such as shops, parks, restaurants, bars, etc. This also can be used to suggest users with same interests in social network services.

\section{Conclusions and Future Work}
\label{sec:ConclusionFutureWork}


Ubiquitous devices such as smart-phones have boosted the interest of researches in location history data such as GPS trajectory records. In this paper, we proposed methods to handle these raw GPS data by applying data mining algorithms to identify meaningful locations. Experimental evaluation was performed in the dataset in order to find meaningful places and similar individuals using the suggested methods. By applying a hybrid method we showed how to extract stay points from users' trajectory data. Furthermore, we were able to find among those extracted stay points those who have more influence in the individuals' routine, here called location points. By analyzing the location points of all the users in a given community we were able to identify the POIs, which are the sets of location points many individuals share in common helping us to understand the users' behavior. Finally, we showed a metric for finding the most similar users in a given community by applying similarity measures over the extracted POIs. This feature is useful to understand the dynamic between individuals in a given community.  The results are quite good when taking into account the precision and the simplicity of the approaches and served as a base to answer the three main questions proposed in the section \ref{sec:Introduction}.

For future work, the next steps include the development of a method to find the patterns of people visiting and leaving different places at different times in a order (weekly basis, daily basis).
Also includes some map matching tasks including external information in order to find the semantic meaning of the individuals' and the collective points of interest or regions of interest. 

\bibliographystyle{splncs04}
\bibliography{biblio} 
\end{document}